# Coronary Heart Disease Diagnosis Based on Improved Ensemble Learning

[1]Kuntoro Adi Nugroho, [2]Noor Akhmad Setiawan, [3]Teguh Bharata Adji
*[1,2,3] Department of Electrical Engineering and Information Technology, Faculty of Engineering, Universitas Gadjah Mada*
*[1] kuntoro_ti08@mail.ugm.ac.id*
*[2] noorwewe@ugm.ac.id*
*[3] adji@ugm.ac.id*

## *Abstract*

*Accurate diagnosis is required before performing proper treatments for coronary heart disease. Machine learning based approaches have been proposed by many researchers to improve the accuracy of coronary heart disease diagnosis. Ensemble learning and cascade generalization are among the methods which can be used to improve the generalization ability of learning algorithm. The objective of this study is to develop heart disease diagnosis method based on ensemble learning and cascade generalization. Cascade generalization method with loose coupling strategy is proposed in this study. C4.5 and RIPPER algorithm were used as meta-level algorithm and Naive Bayes was used as base-level algorithm. Bagging and Random Subspace were evaluated for constructing the ensemble. The hybrid cascade ensemble methods are compared with the learning algorithms in non-ensemble mode and non-cascade mode. The methods are also compared with Rotation Forest. Based on the evaluation result, the hybrid cascade ensemble method demonstrated the best result for the given heart disease diagnosis case. Accuracy and diversity evaluation was performed to analyze the impact of the cascade strategy. Based on the result, the accuracy of the classifiers in the ensemble is increased but the diversity is decreased.*

## 1. Introduction

Cardiovascular disease (CVD) grows along with the growth of economy and living standard [1]. In 2010, cardiovascular disease became the biggest cause of mortality in UK. The number of death caused by CVD is almost 180.000. As reported by British Heart Foundation, approximately 80.000 death was due to coronary heart disease [2]. Cardiovascular diseases, such as coronary heart disease and arrhythmia, are among diseases which endanger human life [1].

The presence of CVD can be detected by some symptoms, such as chest pain and fatigue. Nevertheless, it can not be detected until an attack happened in 50% among reported cases [3]. Diagnosis of cardiovascular disease is an important stage before performing correct treatments. However, performing diagnosis is not an easy task. Therefore, highly skilled physician is required [3], [4].

Statistical and machine learning based approaches have been studied to improve the quality of cardiovascular disease diagnosis. For example, diagnosis characteristic discovery for coronary heart disease with Qi Deficiency Syndrome was proposed by Huihui Zhao, et al. [5]. In this study, t-test based AdaBoost was used to evaluate several biological parameters related to coronary heart disease. Abdel-Motaleb and Akula [6] proposed heart disease diagnosis based on phonogram signals by using Back Propagation and Radial Basis Function Artificial Neural Network. In the study, 94 phonogram signals were separated into 66 training instances, 5 validation instances, and 23 test instances. The performance of RBF network was superior with 98% of accuracy compared to which of Back Propagation network.

Back Propagation network with Levenberg-Marquardt algorithm was proposed by Ming, et al. [7] in Coronary Artery Disease diagnosis. Coronary Heart Disease data from Cleveland Clinic Foundation was used for the experiment. The data comprises 13 conditional attributes and one decision attribute with five values based on the presence and severity of CHD. Two hundreds random samples were selected for training set and the rest were used as testing set. The proposed method demonstrated 99.7% accordance rate. Support Vector Machine was proposed by Zhang, et. al. [8]. Using data set from UCI machine learning repository, 44 samples were taken as training set and 44 samples were taken as testing set. Using PCA with 90% contribution rate and SVM, 84.1% of accuracy was obtained. The proposed methods in the discussed studies demonstrated high prediction rate. However, more proper evaluation methods are required to obtain more confidence in the results, such as k-fold cross validation.

Setiawan, et. al. [9] proposed fuzzy decision support system (FDSS) for coronary heart disease diagnosis based on Rough Set and Fuzzy Set. Rough Set rule induction was used to generate decision rules from data. Rule support and attribute reduction were used as rule filtering to maintain knowledge transparency. To handle imprecise knowledge, fuzzy rules were generated instead of crisp rules. The proposed method successfully demonstrated competitive performance besides maintaining knowledge transparency. In spite of its success, the method required training data to be split into training and testing set for pruning. Thus, reducing the amount of data for training the classifier. Besides that, based on the study, RIPPER, a fast rule induction method, was competitive with FDSS in Cleveland data.

In machine learning and data mining, one of the solution to learn difficult pattern is by using mixture of expert strategy, which is also called ensemble learning. The fundamental concept of ensemble learning is similar with solving a problem by involving multiple experts in a domain. In the classification context, multiple experts are similar to classifiers which are generated from training data [10], [11]. Several advantages of ensemble learning strategy are performing better generalization, handling too little data, and processing large amount of data more efficiently [12].

Besides ensemble learning, cascade generalization strategy [13] could also improve the performance of classification algorithm based on several studies [14], [15]. The general idea of cascade generalization is providing the output of a classifier as an input feature to other classifier in addition to the feature of training data. Instead of using the best pattern recognition method for a certain case, in cascade generalization approach, different pattern recognition strategies are combined [13].

The objective of this study is to improve ensemble learning method for heart disease diagnosis using cascade generalization technique. In this study, cascade classifiers are employed to improve ensemble system for heart disease diagnosis case.

## 2. Introduction to Cascade Generalization and Ensemble Learning

### 2.1. Cascade Generalization

Cascade generalization could be implemented in loose coupled way or tight coupled way [13]. Loose coupling strategy works by implementing classifier in sequential manner which the output of a classifier becomes the input for the following classifier. Tight coupling strategy works in divide and conquer based algorithm by implementing loose coupling strategy locally.

Cascade generalization is based on meta-learning strategy. In this way, classifier works in base level and meta-level. Classifier in base level supplies its output to the classifier in meta level. Thus, meta classifier makes the final decision not only based on input feature, but also based on the output of the base classifier.

Gama and Brazdil [13] provided the formulation of cascade generalization strategy. Let $\tau$ to denote a learning algorithm which induce a classification model, such as C4.5 or Naive Bayes. A classification model $\tau(D)$ is built by implementing learning algorithm $\tau$ to training data D.

Concatenating operation is one of basic operation in cascade generalization, which is also called constructive operator $\Phi$ . The attribute of an instance is extended using probability class distribution obtained by a classification model. A constructive operator with two parameter $(x, \tau(D))$ indicates concatenation of class probability output obtained from model $\tau(D)$ to instance $x$ .

Gama and Brazdil formulates constructive operator when applied to a data set as in Equation 1. A new data set D'' is produced by concatenating data set D' with class probability output predicted by model $\tau(D)$ applied to data set D'. Operator $A(\tau(D), D')$ indicates the classification of data set D' using model $\tau(D)$.

$$D'' = \Phi\left(D', A\left(\tau\left(D\right), D'\right)\right) \tag{1}$$

Cascading can be done in a number of level. Training data on i[th] level is denoted as $train_i$ while test data on i[th] level is denoted as $test_i$. Consider a two level cascading. The base classifier works in level 1 and the meta classifier works on level 2. The original data set which is used to train the base classifier works on level 0, denoted as $train_0$. The learning algorithm that works in meta level uses data on the 1[st] level which are $train_1$ and $test_1$.

Let $\tau_2 \Delta \tau_1$ to denote cascade generalization of learning algorithm $\tau_2$ after $\tau_1$. Thus, $\tau_2$ works in meta level and $\tau_1$ works in base level. The cascade generalization of the two classifiers can be written as in Equation 2.

$$A\left(\tau_2\left(train_1\right), test_1\right) \tag{2}$$

Where $train_1$ and $test_1$ are formulated as in Equation 3.

$$\begin{gathered} train_1 = \Phi\left(train_{0,} A\left(\tau\left(train_0\right), train_0\right)\right) \\ test_1 = \Phi\left(test_{0,} A\left(\tau\left(train_0\right), test_0\right)\right) \end{gathered} \tag{3}$$

## 2.2. Ensemble Learning Algorithm

Ensemble learning works by constructing multiple model for a given data sets. Several algorithms for constructing ensemble have been proposed. These methods aim to construct diverse models but still accurate.

Besides ensemble construction, decision fusion procedure must be defined. Decision fusion procedure is used to obtain the final decision by considering the decision of the classifiers in the ensemble. There are several known decision fusion procedures, such as majority voting, weighted voting, and stacked generalization [11].

Figure 1 illustrates an example of ensemble system. Let the term base classifier in ensemble context refer to the classifiers which comprise an ensemble. The ensemble consists of four base classifiers, $\tau_0$, $\tau_1$, $\tau_2$, and $\tau_3$, which can be in any form, such as decision tree or artificial neural network. Each of the classifier process a given test instance to obtain decision. After that, the decisions given by all of the classifiers are processed by decision fusion subsystem to obtain final decision.

There are some general approach in constructing classifier ensemble [16], [17]:

- Instance perturbation: altering training instance to construct diverse classifiers.
- Feature perturbation: altering feature set to construct diverse classifiers.
- Classifier perturbation: using different learning algorithms to construct diverse classifiers.
- Hybrid technique: combining different types of perturbation.

Bagging and Random Subspace methods are ensemble construction methods which based on instance perturbation and feature perturbation respectively.

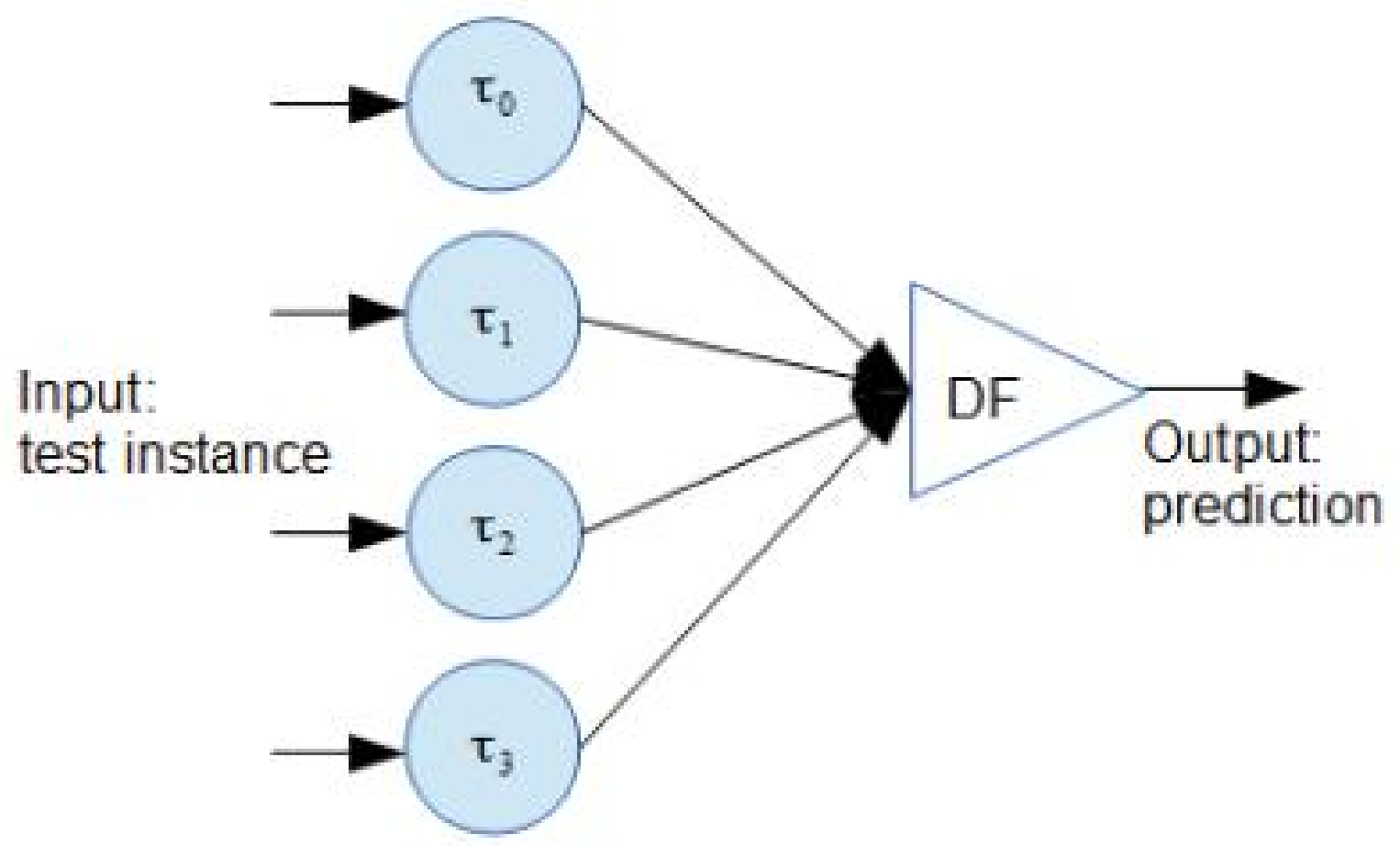

**Figure 1.** Ensemble system composed of 4 classifiers: an illustration

### 2.2.1. Bagging

Bagging [18] can be used to construct multiple different classifier by using different training set for each classifier. Each training set for a classifier is created by taking bootstrap sample of the original training data. Bagging can be advantageous when used for unstable classifier, such as decision tree or artificial neural network. The pseudo-code for Bagging is given in Algorithm 1 [11], [18].

**Algorithm 1**: Bagging
**Input:**
L = learning algorithm
D = training set
n = number of generated classifier
Output:
T = a set of classifier
Begin
for i = 1 to n
    $S_i$ := bootstrap sample from D
    Construct classifier $T_i$ by applying learning algorithm L on $S_i$
end for
return T
End

To obtain the final decision for a given test instance, some procedure can be used, such as majority voting [11]. In majority voting, the final decision of the ensemble is the class which gains the majority support from the classifiers.

### 2.2.2. Random Subspace

While Bagging constructs ensemble using training instance bootstrapping, Random Subspace [19] constructs ensemble based on random selection of feature space. In this method, full training data is used to train classifier. Using feature subset can be advantageous as reducing data dimension. Moreover, the required training time for using this method can be made more efficiently as it is a “parallel learning algorithm” [19].

# 3. Methodology

## 3.1. The Proposed Algorithms

Many statistical and data mining approach have been proposed for heart disease diagnosis, such as artificial neural network [6], [7] and rule induction based method [9]. Ensemble learning is suggested for improving generalization ability of learning algorithm. In order for ensemble to perform well, the base classifiers of an ensemble should be both accurate and diverse [20]. In this study, cascade generalization is proposed as a method to increase the base classifier's accuracy in the ensemble. Instead of using learning algorithm in normal way to construct ensemble, cascade learning algorithm is used to construct classifiers in the ensemble.

Cascade classifier with loose coupling strategy is used in this study. The cascade strategy uses two level cascading, thus two classifiers are employed. One classifier is used in base level and the other is used in meta level.

Three types of learning algorithms are evaluated in this study, namely Naive Bayes, C4.5, and RIPPER. C4.5 is one of decision tree induction algorithm developed as improvement from ID3 [21]. The improvements include missing value and numerical attribute handling. RIPPER (Repeated Incremental Pruning to Produce Error Reduction) is one of rule induction methods which introduces several improvements to IREP in terms of rule value metric, stopping condition for rule generation, and rule optimization [22]. Naive Bayes [23] is a classification learning algorithm which is based on Bayes's theorem. In this method, all attributes in the data set are assumed independent.

Naive Bayes is implemented in base-level of the cascade. Naive Bayes had previously used in the study by Gama and Brazdil [13] as base classifier. Naive Bayes assumes that all attributes are independent. To seize attribute correlation problem, attribute selection is applied to the data before used to train Naive Bayes classifier. Genetic Algorithm is selected as attribute selection search method [24], while correlation based feature subset evaluator is used for feature evaluation [25].

C4.5 and RIPPER (Repeated Incremental Pruning to Produce Error Reduction) are proposed as learning algorithm in the meta-level of the cascade. Both are learning algorithm with different knowledge representation. RIPPER constructs decision rule while C4.5 constructs decision tree. RIPPER had previously evaluated as meta-classifier in the study of CGen-SVM, an ensemble in which contain one cascade classifier [26]. Evaluated on five data sets, RIPPER demonstrated the second best result with only small difference to the best meta-classifier. C4.5 had also used as meta level classifier in some previous studies [14], [15] and demonstrated increase in performance.

The problem of heart disease diagnosis in this study is a binary classification task. The problem is to discern whether a given case is diseased or not-diseased. If the probability of diseased case is p, then the probability of not-diseased case is (1-p). Thus, in this study, only one attribute will be added to training data for meta classifier training.

## 3.2. Several Measurements Indicator

### 3.2.1. Accuracy

Accuracy measures the percentage of correct predictions for a given test data [27]. Equation 4 denotes the formulation of accuracy.

$$Accuracy = \frac{Numberofcorrectlyclassifierinstance}{Numberoftestinstance} \tag{4}$$

### 3.2.2. Sensitivity and Specificity

For a binary class classification, such as diagnosing whether the presence of a certain disease is positive or negative, the performance measure for positive and negative class can be defined. Sensitivity or true positive rate is a measure of accuracy for positive class while specificity or true negative rate is a measure of accuracy for negative class [28].

Sensitivity and specificity can be expressed as in Equation 5 and 6 respectively. TP denotes the number of correctly classified positive class while TN denotes the number of correctly negative class. P and N refer to the number of positive and negative instances respectively.

True Negative and True Positive rate represents the ability of the method in predicting negative class and positive class. In this case, negative class indicates no presence of heart disease and positive class indicates the presence of heart disease. When a classifier produced by learning algorithm possessed high True Negative rate but low True Positive rate, the classifier predicts negative class more. In extreme case, a classifier will obtain 100% true negative rate but 0% true positive rate when the classifier always predicts negative case.

$$Sensitivity = True\ Positive\ Rate = \frac{TP}{P} \tag{5}$$

$$Specificity = True\ Negative\ Rate = \frac{TN}{N} \tag{6}$$

### 3.2.3. Receiver Operating Characteristic Area Under Curve (ROC AUC)

For a given diagnosis which is measured in probability, a given threshold must be defined to determine the class label prediction. Hence, performance measures such as sensitivity and specificity depend on the given decision threshold [29].

ROC AUC can be used to measure diagnosis performance independent from decision threshold as it measures the trade-off between true positive rate and true negative rate for every possible threshold [29]. ROC graph contains the plot of sensitivity and (1 – specificity) for every possible threshold. ROC AUC is measured by computing the area below ROC graph which maximum value is one.

### 3.2.4. Accuracy and Diversity

Both accuracy and diversity are factors which affect the quality of an ensemble. High accuracy and high diversity are required to construct good ensemble [20]. In the context of accuracy and diversity analysis, accuracy can be measured as the average accuracy of the classifiers in an ensemble.

Diversity can be seen as heterogeneity of the classifiers in the ensemble. Several measures can be used to evaluate diversity, but the best measure is unclear. Kuncheva and Whitaker suggested Q statistics, but the underlying reason for the selection is speculative [30]. Beside that, based on the study, the evaluated diversity measures are strongly correlated between themselves.

In this study, Kohavi Wolpert variance is used to measure diversity [31]. Kohavi Wolpert variance is a non-pairwise measure in which the diversity is not measured for each pair of classifier. Kohavi Wolpert variance is proportional to diversity, which means the higher the variance, the higher the diversity.

## 3.3. Tools

In this study, the following software libraries are used

- Weka [32] API
- Genetic Search API [33]

## 3.4. Data

Cleveland heart disease data from UCI Machine Learning Repository [34] is used in this study. This data was collected in Cleveland Clinic Foundation [35]. The purpose of this data is to discriminate patient with coronary heart disease.

The data comprises 303 instances, 13 attributes, and one decision attribute. The decision attribute has five values. The value ranges from 0 to 4. Zero indicates no presence of disease and 1-4 indicates the presence of disease. However, only two class labels are considered in this study. One label indicate no presence and the other indicates the presence of the disease. Thus, there are 164 cases with no presence of disease and 139 cases with presence of disease.

The attributes are described in Table 1. Among the thirteen attributes, seven attributes are categorical attribute and the rest are numeric.

## 3.5. Experimental Methodology

The following thirteen classification methods are evaluated in this study

- C4.5, Bagging of C4.5, Random Subspace of C4.5
- RIPPER, Bagging of RIPPER, Random Subspace of RIPPER
- Cascade of C4.5, Bagging of cascade C4.5, Random Subspace of cascade C4.5
- Cascade of RIPPER, Bagging of cascade RIPPER, Random Subspace of cascade RIPPER
- Naive Bayes
- For comparison, Rotation Forest ensemble [36] is also evaluated using C4.5 as base classifier.

The following are the configuration settings for the evaluation

- All the evaluated ensemble method comprises thirty classifiers.
- Twenty five runs of 10-fold cross validation are performed for the evaluation.
- The performance evaluation metrics are accuracy, ROC Area Under Curve, Sensitivity, and Specificity.
- Accuracy and diversity analysis are performed to understand the impact of the cascade strategy to the ensemble methods.

**Table 1.** Cleveland heart data attributes

| Attribute name | Description | Type |
|---|---|---|
| age | Age of patient in years | numeric |
| sex | Sex (female = 0, male = 1) | nominal |
| cp | Chest pain type (typical angina = 1, atypical angina = 2, non-anginal pain = 3, asymptomatic = 4) | nominal |
| trestbps | Resting blood pressure (mm Hg) on admission to the hospital | numeric |
| chol | Serum cholesterol (mg/dl) | numeric |
| fbs | If fasting blood sugar > 120mg/dl (yes = 1, no = 0) | nominal |
| restecg | Resting electrocardiographic result (normal = 0, having ST-T wave abnormality = 1, showing probable or definite left ventricular hypertrophy by Estes' criteria = 2) | nominal |
| thalach | Maximum heart rate achieved | numeric |
| exang | Exercise induced angina (no = 0, yes = 1) | nominal |
| oldpeak | ST depression induced by exercise relative to rest | numeric |
| slope | The slope of the peak exercise ST segment (up-sloping = 1, flat = 2, down-sloping = 3) | nominal |
| ca | Number of major vessels colored by flourosopy (0-3) | numeric |
| thal | Normal = 3, Fixed defect = 6, Reversable defect = 7 | nominal |
| num | Diagnosis of heart disease (0 indicates < 50% diameter narrowing while 1 indicates > 50% diameter narrowing) | nominal |

## 4. Result and Discussion

Table 2 depicted the summary of the evaluation result. The first column indicates the evaluated methods. The first row indicates the evaluated performance metrics. The cascade version of classifier is begin with "C", for example, C-RPR indicates the cascaded version of the RIPPER classifier. Bagging and Random Subspace version of the classifiers are indicated with "Bg" and "RS" respectively. The highest result is printed in bold. Three highest results per indicator are printed in blue background.

The accuracy, ROC, TN rate, and TP rate are depicted in Figure 2 and Figure 3. Figure 2 illustrates the accuracy and the ROC AUC and Figure 3 illustrates the TN and TP rate in bar chart for all methods. All of the values are normalized in 0 to 100.

The result demonstrated that ensemble with cascade generalization strategy could improve the performance of learning algorithm for the given heart disease diagnosis case. Normally, C4.5 achieved only 78.204% accuracy, 0.79332 ROC AUC, 0.83951 true negative rate, and 0.71424 true positive rate. When C4.5 was implemented using the proposed cascade strategy, there is significance increase in the accuracy, the ROC AUC, the TN rate, and the TP rate. The improvements are even higher when the cascade C4.5 was used in Bagging or Random Subspace. The ensemble version of cascade C4.5 also demonstrated better result compared to the ensemble version of C4.5.

In terms of accuracy, Bagging of cascade C4.5 achieved the best result with slight difference compared to the second, the third, and the fourth best method. The superiority in accuracy indicates that Bagging of cascade C4.5 demonstrated the best correct diagnosis rate among other methods. Naive Bayes is the best in terms of ROC, followed by Random Subspace of cascade C4.5 with slight difference. High ROC Area Under Curve indicates that Naive Bayes and Random Subspace of cascade C4.5 possessed good trade-off in sensitivity and specificity.

**Table 2.** Result summary

| | **Accuracy (%)** | **ROC AUC** | **TN Rate** | **TP Rate** |
|---|---|---|---|---|
| RPR (RIPPER) | 78.20462 | 0.79332 | 0.83951 | 0.71424 |
| Bg – RPR (Bagging – RIPPER) | 82.29703 | 0.89404 | 0.87732 | 0.75885 |
| Rs – RPR (Random Subspace – RIPPER) | 82.37624 | 0.89590 | 0.89561 | 0.73899 |
| C4.5 (C4.5) | 76.15842 | 0.76599 | 0.79707 | 0.71971 |
| Bg – C4.5 (Bagging – C4.5) | 79.59076 | 0.88351 | 0.82122 | 0.76604 |
| Rs – C4.5 (Random Subspace – C4.5) | 81.57096 | 0.89166 | 0.86244 | 0.76058 |
| C_RPR (Cascade RIPPER) | 82.15182 | 0.80901 | 0.88268 | 0.74935 |
| Bg – C_RPR (Bagging – Cascade RIPPER) | 83.41914 | 0.89108 | 0.87829 | 0.78216 |
| Rs – C_RPR (Random Subspace – Cascade RIPPER) | 83.47195 | 0.89398 | 0.89293 | 0.76604 |
| C_C4.5 (Cascade C4.5) | 80.56766 | 0.80153 | 0.84171 | 0.76317 |
| Bg – C_4.5 (Bagging – Cascade C4.5) | 83.57756 | 0.89603 | 0.86878 | 0.79683 |
| Rs – C_C4.5 (Random Subspace – Cascade C4.5) | 83.49835 | 0.90079 | 0.88829 | 0.77209 |
| RS (Rotation Forest) | 82.16502 | 0.89411 | 0.85268 | 0.78504 |
| NB (Naive Bayes) | 82.91749 | 0.90204 | 0.86146 | 0.79108 |

There is no single method which performed well in both true negative rate and true positive rate. Random Subspace of RIPPER demonstrated the best True Negative rate, but low True Positive rate compared to other methods. Bagging of cascade C4.5 demonstrated the best performance in True Positive rate. However, this method still demonstrated moderate performance in True Negative rate. Thus, by considering all the four metrics, Bagging of cascade C4.5 performed better than Random Subspace of RIPPER.

Based on the result from this study, Bagging and Random Subspace of cascade C4.5 demonstrated high performance in three of four metrics compared to other methods. The result indicates that ensemble of cascade classifier could improve C4.5 for the heart disease diagnosis case.

Cascade generalization brings performance improvement impact on Bagging and Random Subspace with C4.5. To understand the reason for the improvement, accuracy and diversity analysis were performed. The ensemble methods with C4.5 and cascade C4.5 are evaluated. The results are summarized in Table 3. The average accuracy of the classifiers constructed with Bagging and Random Subspace using C4.5 are 74.5404% and 75.4762% respectively. The cascade strategy increase the average accuracy for both Bagging (78.2409%) and Random Subspace (77.4704%). However, in spite of the increase in accuracy, the diversity of the methods are decreased. For instance, the diversity of Bagging with C4.5 is 0.1166 and the diversity of Bagging with cascade C4.5 is 0.0922.

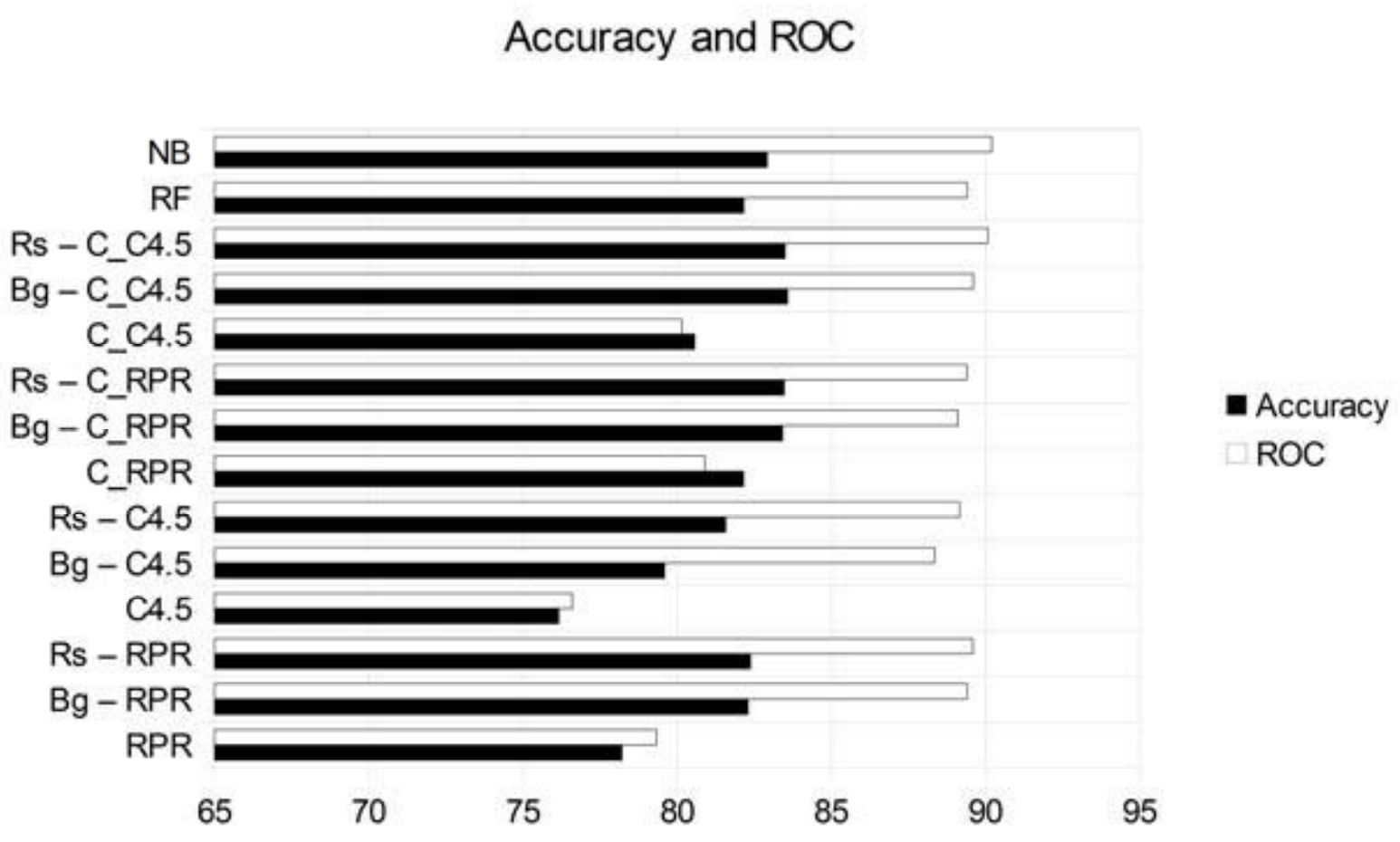

**Figure 2.** Accuracy and ROC results

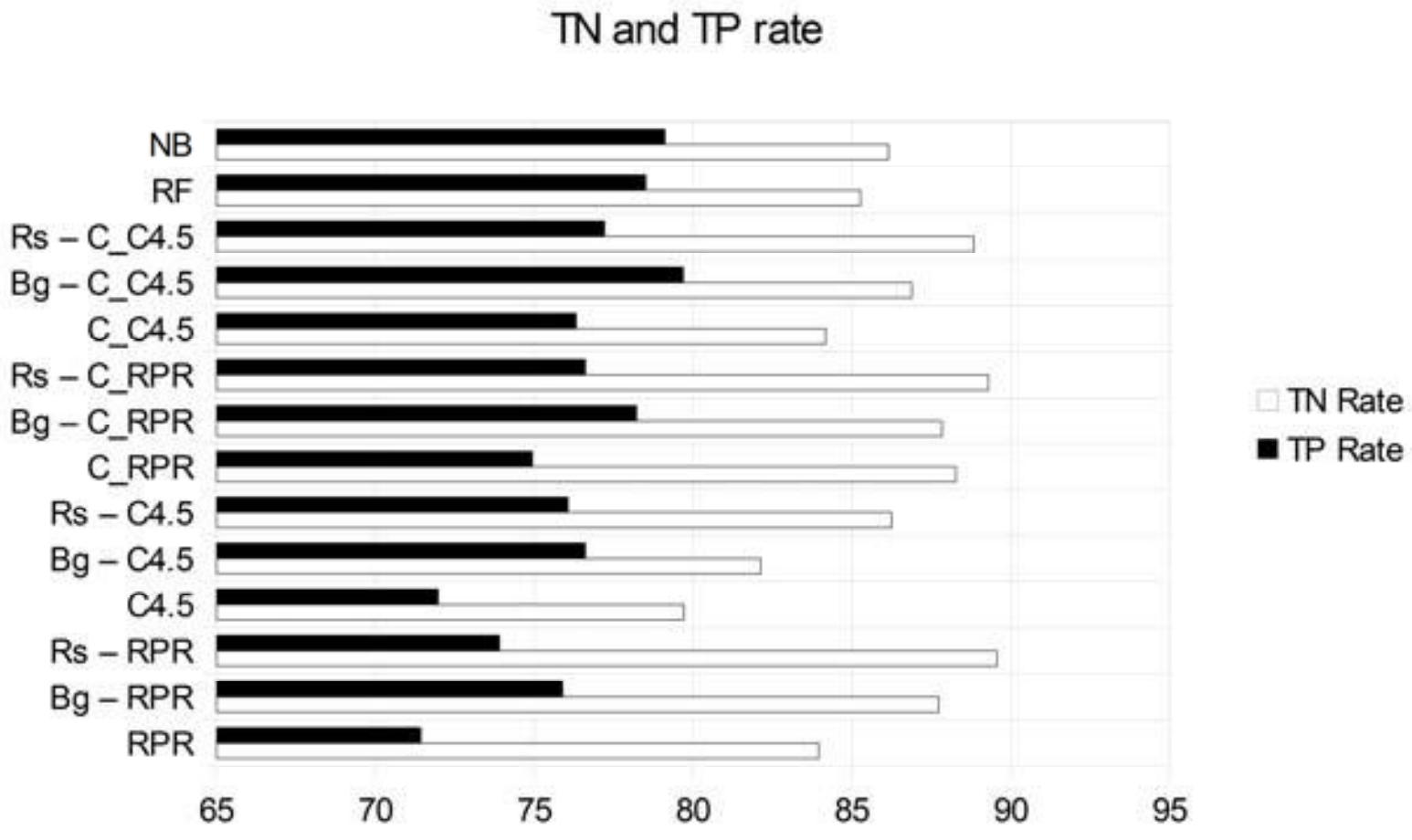

**Figure 3.** TN and TP rate

**Table 3.** Accuracy and diversity analysis

| | Diversity | Accuracy |
|---|---|---|
| Bagging – C4.5 | 0.1166 | 74.5404 |
| Random Subspace – C4.5 | 0.1135 | 75.4762 |
| Bagging – C_C4.5 | 0.0922 | 78.2409 |
| Random Subspace – C_C4.5 | 0.1019 | 77.4704 |

Based on the accuracy and diversity evaluation result, the cascade procedure improves the performance of ensemble method by increasing the accuracy but decreasing the diversity at the same time, although this result is not generalizable. In other words, whether ensemble with high accuracy but low diversity will always outperformed ensemble with lower accuracy but higher diversity is uncertain. This result does not implies that to improve the performance of an ensemble, the accuracy of the classifiers must be set higher and the diversity of the classifiers must be set lower. Further study is required to understand the relation among base classifier accuracy, base classifier diversity, and the accuracy of an ensemble.

## 5. Conclusion

Ensemble of cascade classifiers for heart disease diagnosis has been proposed in this study. The proposed method was evaluated on heart disease diagnosis case based on data from Cleveland Clinic Foundation. C4.5 and RIPPER were implemented in the ensemble of cascade strategy. Naive Bayes with feature selection was employed to extend the input feature of the meta-classifier in the cascade. The result demonstrated that Bagging and Random Subspace of cascade C4.5 demonstrated high performance for the given heart disease diagnosis case. The proposed cascade strategy increase the average accuracy but decrease the diversity of the ensemble at the same time.